\documentclass{article}

\usepackage[T1]{fontenc}
\usepackage{tgtermes}
\usepackage[utf8]{inputenc}

\usepackage{amssymb}
\newcommand{\mathbold}[1]{\ensuremath{\boldsymbol{\mathbf{#1}}}}

\usepackage{scalefnt,letltxmacro}
\LetLtxMacro{\oldtextsc}{\textsc}
\renewcommand{\textsc}[1]{\oldtextsc{\scalefont{1.10}#1}}
\usepackage[ttdefault=true]{AnonymousPro}

\usepackage{amsmath}
\usepackage{amsthm}

\usepackage[english]{babel}
\usepackage[parfill]{parskip}
\usepackage{afterpage}
\usepackage{enumitem}
\usepackage{framed}
\usepackage{xspace}

\usepackage{lineno}

\usepackage{ragged2e}

\newcounter{parcount}

\DeclareRobustCommand{\parhead}[1]{\textbf{#1}~}
\usepackage[usenames,dvipsnames]{xcolor}
\definecolor{shadecolor}{gray}{0.9}

\usepackage{graphicx}
\usepackage[labelfont=bf]{caption}
\usepackage[format=hang]{subcaption}

\usepackage{booktabs, array}

\usepackage[algoruled]{algorithm2e}
\usepackage{listings}
\usepackage{fancyvrb}
\fvset{fontsize=\small}

\usepackage[numbers]{natbib}
\usepackage[colorlinks,linktoc=all]{hyperref}
\usepackage[all]{hypcap}
\hypersetup{citecolor=MidnightBlue}
\hypersetup{linkcolor=MidnightBlue}
\hypersetup{urlcolor=MidnightBlue}

\usepackage[nameinlink]{cleveref}
\crefname{section}{§}{§§}
\Crefname{section}{§}{§§}
\creflabelformat{equation}{#2\textup{#1}#3}  

\newcommand{\myeqp}[1]{\hyperref[eq:#1]{Eq.\ref*{eq:#1}}}
\newcommand{\mysec}[1]{\hyperref[sec:#1]{Section~\ref*{sec:#1}}}
\newcommand{\mysub}[1]{\hyperref[sub:#1]{Section~\ref*{sub:#1}}}
\newcommand{\mytable}[1]{\hyperref[table:#1]{Table~\ref*{table:#1}}}
\newcommand{\myfig}[1]{\hyperref[fig:#1]{Fig.~\ref*{fig:#1}}}
\newcommand{\myappendix}[1]{\hyperref[appendix:#1]{Appendix~\ref*{appendix:#1}}}
\newcommand{\myalg}[1]{\hyperref[alg:#1]{Algorithm~\ref*{alg:#1}}}
\newcommand{\mytheorem}[1]{\hyperref[theorem:#1]{Theorem~\ref*{theorem:#1}}}
\newcommand{\myfootnote}[1]{\hyperref[footnote:#1]{Footnote~\ref*{footnote:#1}}}

\usepackage
[acronym,smallcaps,nowarn,section,nonumberlist]{glossaries}
\glsdisablehyper{}

\definecolor{strings}{rgb}{.624,.251,.259}
\definecolor{keywords}{rgb}{.224,.451,.686}
\definecolor{comment}{rgb}{.322,.451,.322}

\lstdefinelanguage{python}{
  keywords=[3]{Normal, Bernoulli, Beta, Categorical, Dirichlet,
  Exponential, MultivariateNormalFull, RandomVariable,
  DirichletProcess, Empirical, PointMass, Gamma,
  MAP, Inference, KLqp, HMC, SGLD, KLpq,
  VariationalInference, MonteCarlo, ConjugateInference, GANInference,
  rnn_cell, dirichlet_process, cond, body, generative_network,
  discriminative_network, Dense,
  evaluate, ppc, copy, dot, get_session},
  morecomment=[l]{\#},
  morecomment=[s]{"""}{"""},
  morestring=[b]',
  morestring=[b]",
  alsoletter={<>=-+/*},
  sensitive=true
}

\renewcommand{\texttt}[1]{\lstinline[basicstyle=\fontsize{9pt}{9.25pt}\selectfont\ttfamily]{#1}}

\newacronym{KL}{kl}{Kullback-Leibler}
\newacronym{ELBO}{elbo}{\emph{evidence lower bound}}
\newacronym{GAN}{gan}{generative adversarial network}
\newacronym{ABC}{abc}{approximate Bayesian computation}
\newacronym{HIM}{him}{hierarchical implicit model}
\newacronym{DIM}{dim}{deep implicit model}
\newacronym{RNN}{rnn}{recurrent neural network}
\newacronym{LFVI}{lfvi}{likelihood-free variational inference}

\newcommand{\g}{\,|\,}

\renewcommand{\d}[1]{\ensuremath{\operatorname{d}\!{#1}}}
\newcommand{\nestedmathbold}[1]{{\mathbold{#1}}}

\newcommand{\mbw}{\nestedmathbold{w}}
\newcommand{\mbx}{\nestedmathbold{x}}
\newcommand{\mby}{\nestedmathbold{y}}
\newcommand{\mbz}{\nestedmathbold{z}}

\newcommand{\mbX}{\nestedmathbold{X}}

\newcommand{\mbbeta}{\nestedmathbold{\beta}}
\newcommand{\mbdelta}{\nestedmathbold{\delta}}
\newcommand{\mbepsilon}{\nestedmathbold{\epsilon}}

\newcommand{\mblambda}{\nestedmathbold{\lambda}}

\newcommand{\mbphi}{\nestedmathbold{\phi}}

\newcommand{\mbtheta}{\nestedmathbold{\theta}}

\newcommand{\cD}{\mathcal{D}}
\newcommand{\cL}{\mathcal{L}}
\newcommand{\cP}{\mathcal{P}}
\newcommand{\cQ}{\mathcal{Q}}
\newcommand{\cF}{\mathcal{F}}

\newcommand{\E}{\mathbb{E}}

\newcommand{\Tglobal}{T_\textrm{global}}
\newcommand{\Tlocal}{T_\textrm{local}}
\newcommand{\deltaglobal}{\mbdelta_\textrm{global}}
\newcommand{\deltalocal}{\mbdelta_n}
\newcommand{\nglobal}{s(\cdot)}
\newcommand{\nlocal}{s(\cdot)}

\usepackage{tikz}

\usetikzlibrary{bayesnet} 

\pgfdeclarelayer{edgelayer}
\pgfdeclarelayer{nodelayer}
\pgfsetlayers{edgelayer,nodelayer,main}

\definecolor{hexcolor0xbfbfbf}{rgb}{0.749,0.749,0.749}

\tikzset{>=latex}
\tikzstyle{none}   = [inner sep=0pt]
\tikzstyle{line}   = [ -, thick, shorten <=1pt, shorten >=1pt ]
\tikzstyle{arrow}  = [ ->, thick, shorten <=1pt, shorten >=1pt ]
\tikzstyle{ardash} = [ dashed, ->, thick, shorten <=1pt, shorten >=1pt ]

\tikzstyle{empty}=[circle,opacity=0.0,text opacity=1.0,inner sep=0pt]
\tikzstyle{box}=[rectangle,fill=White,draw=Black]
\tikzstyle{filled}=[circle,thick,fill=hexcolor0xbfbfbf,draw=Black]
\tikzstyle{hollow}=[circle,thick,fill=White,draw=Black]
\tikzstyle{param}=[rectangle,fill=Black,draw=Black,inner sep=0pt,minimum width=4pt,minimum height=4pt]
\tikzstyle{paramhollow}=[rectangle,thick,fill=White,draw=Black,inner sep=0pt,minimum
width=4pt,minimum height=4pt]

\usepackage{pgfplots}                               
\pgfplotsset{compat=newest}
\pgfplotsset{plot coordinates/math parser=false}
\newlength\figureheight
\newlength\figurewidth
\newlength\figureheightsmall
\newlength\figurewidthsmall
\definecolor{POSTcolor}{rgb}{0.48, 0.20, 0.58} 
\definecolor{Qcolor}{rgb}{0.00, 0.53, 0.22} 

\usepackage[final]{nips_2017}
\title{%
Hierarchical Implicit Models and \\ Likelihood-Free Variational Inference
}

\author{
Dustin Tran \\
Columbia University \\
\And
Rajesh Ranganath \\
Princeton University \\
\And
David M. Blei \\
Columbia University \\
}

\begin{document}

\maketitle

\begin{abstract}
Implicit probabilistic models are a flexible class of models
defined by a simulation process for data.
They form the basis for theories which encompass our understanding
of the physical world. Despite this fundamental nature, the use of
implicit models remains limited due to challenges in specifying complex latent
structure in them, and in performing inferences in such models with large data sets.
In this paper, we first introduce \glsreset{HIM}\emph{\glspl{HIM}}.
\glspl{HIM} combine the idea of implicit densities with hierarchical
Bayesian modeling, thereby defining models via simulators of data
with rich hidden structure. Next, we develop
\glsreset{LFVI}\emph{\gls{LFVI}}, a scalable variational inference
algorithm for \glspl{HIM}.
 Key to
\gls{LFVI} is
specifying a variational family that is also implicit. This matches
the model's flexibility and allows for accurate approximation of the
posterior.
We
demonstrate diverse applications: a large-scale physical simulator for
predator-prey populations in ecology; a Bayesian \acrlong{GAN} for
discrete data; and a deep implicit model for text generation.
\end{abstract}

\section{Introduction}

Consider a model of coin tosses.  With probabilistic models, one
typically posits a latent probability, and supposes each toss is a
Bernoulli outcome given this probability
\citep{murphy2012machine,gelman2013bayesian}.  After observing a
collection of coin tosses, Bayesian analysis lets us describe our
inferences about the probability.

However, we know from the laws of physics that the outcome of a coin
toss is fully determined by its initial conditions (say, the impulse
and angle of flip)
\citep{keller1986probability,diaconis2007dynamical}.  Therefore a coin
toss' randomness does not originate from a latent probability but in
noisy initial parameters.  This alternative model incorporates the
physical system, better capturing the generative process.  Furthermore
the model is \emph{implicit}, also known as a simulator: we can sample
data from its generative process, but we may not have access to
calculate its density \citep{diggle1984monte,hartig2011statistical}.

Coin tosses are simple, but they serve as a building block for complex
implicit models.  These models, which capture the laws and theories of
real-world physical systems, pervade fields such as population
genetics \citep{pritchard1999population}, statistical physics
\citep{anelli2008totem}, and ecology \citep{beaumont2010approximate};
they underlie structural equation models in economics and causality
\citep{pearl2000causality}; and they connect deeply to \glspl{GAN}
\citep{goodfellow2014generative}, which use neural networks to specify
a flexible implicit density~\citep{mohamed2016learning}.

Unfortunately, implicit models, including \glspl{GAN}, have seen
limited success outside specific domains.  There are two reasons.
First, it is unknown how to design implicit models for more general
applications, exposing rich latent structure such as priors,
hierarchies, and sequences.  Second, existing methods for inferring
latent structure in implicit models do not sufficiently scale to
high-dimensional or large data sets.  In this paper, we design a new
class of implicit models and we develop a new algorithm for accurate
and scalable inference.

For modeling, \Cref{sec:model} describes \emph{hierarchical
  implicit models}, a class of Bayesian hierarchical models which only
assume a process that generates samples.  This class encompasses both
simulators in the classical literature and those employed in
\glspl{GAN}.  For example, we specify a Bayesian \gls{GAN}, where
we place a prior on its parameters.  The Bayesian perspective allows
\glspl{GAN} to quantify uncertainty and improve data efficiency.  We
can also apply them to discrete data; this setting is not possible
with traditional estimation algorithms for \glspl{GAN}~\citep{kusner2016gans}.

For inference, \Cref{sec:inference} develops
\glsreset{LFVI}\emph{\gls{LFVI}}, which combines variational inference
with density ratio estimation
\citep{sugiyama2012density,mohamed2016learning}.  Variational
inference posits a family of distributions over latent variables and
then optimizes to find the member closest to the posterior
\citep{jordan1999variational}.  Traditional approaches require a
likelihood-based model and use crude approximations, employing a
simple approximating family for fast computation.  \gls{LFVI} expands
variational inference to implicit models and enables accurate
variational approximations with implicit variational families:
\gls{LFVI} does not require the variational density to be tractable.
Further, unlike previous Bayesian methods for implicit models,
\gls{LFVI} scales to millions of data points with stochastic
optimization.

This work has diverse applications.  First, we analyze a classical
problem from the \gls{ABC} literature, where the model simulates an
ecological system \citep{beaumont2010approximate}. We analyze 100,000
time series which is not possible with traditional methods.  Second,
we analyze a Bayesian \gls{GAN}, which is a \gls{GAN} with a prior
over its weights. Bayesian \glspl{GAN} outperform corresponding Bayesian neural networks with known likelihoods on several classification tasks.
Third, we show how injecting noise into hidden units of
\acrlongpl{RNN} corresponds to a deep implicit model for flexible
sequence generation.

\vspace{-1.5ex}
\paragraph{Related Work.}
This paper connects closely to three lines of work.  The first is
Bayesian inference for implicit models, known in the statistics
literature as \glsreset{ABC}\gls{ABC}
\citep{beaumont2010approximate,marin2012approximate}.  \gls{ABC} steps
around the intractable likelihood by applying summary statistics to
measure the closeness of simulated samples to real observations. While
successful in many domains, \gls{ABC} has shortcomings. First,
the results generated by \gls{ABC} depend heavily on the chosen
summary statistics and the closeness measure. Second, as the
dimensionality grows, closeness becomes harder to achieve. This is the
classic curse of dimensionality.

\vspace{-0.5ex} The second is \glspl{GAN}
\citep{goodfellow2014generative}.  \glspl{GAN} have seen much interest
since their conception, providing an efficient method for estimation
in neural network-based simulators.  \citet{larsen2016autoencoding}
propose a hybrid of variational methods and \glspl{GAN} for improved
reconstruction.  \citet{chen2016infogan} apply information penalties
to disentangle factors of variation.
\citet{donahue2017adversarial,dumoulin2017adversarially} propose to
match on an augmented space, simultaneously training the model and an
inverse mapping from data to noise.  Unlike any of the above, we
develop models with explicit priors on latent
variables, hierarchies, and sequences, and we
generalize \glspl{GAN} to perform Bayesian inference.

\vspace{-0.5ex}
The final thread is variational inference with expressive approximations
\citep{rezende2015variational,salimans2015markov,tran2015copula}.
The idea of casting the design of variational families as a modeling
problem was proposed in
\citet{ranganath2016hierarchical}.
Further advances have analyzed variational programs
\citep{ranganath2016operator}---a family of approximations which only
requires a process returning samples---and which has seen further
interest \citep{liu2016two}.
Implicit-like variational
approximations have also appeared in auto-encoder frameworks
\citep{makhzani2015adversarial,mescheder2017adversarial}
and message passing \citep{karaletsos2016adversarial}.
We build on variational programs for inferring implicit models.

\vspace{-0.5ex}

\section{Hierarchical Implicit Models}
\label{sec:model}

\begin{figure}[tb]
\centering
\begin{subfigure}{0.4\columnwidth}
  \centering
  \begin{tikzpicture}

  \node[obs, minimum size=0.5cm] (x)    {} ;
  \node[latent, above=of x, minimum size=0.5cm] (z1)    {} ;
  \node[latent, left=0.5cm of z1, minimum size=0.5cm] (beta1)    {} ;

  \node[right=of x, xshift=-1cm] (xlabel) {$\mbx_n$};
  \node[right=0.05cm of z1] (z1label) {$\mbz_{n}$};
  \node[left=0.04cm of beta1] (beta1label) {$\beta$};

  \edge{z1}{x};
  \edge{beta1}{z1};
  \edge{beta1}{x};

  \plate[inner sep=0.2cm,xshift=-0.2cm,
    label={[xshift=-15pt,yshift=15pt]south east:$N$}] {plate1} {
    (x)(xlabel)(z1)(z1label)
  } {};

\end{tikzpicture}
  \label{sub:hierarchical_model}
\end{subfigure}%
\begin{subfigure}{0.6\columnwidth}
  \centering
  \begin{tikzpicture}

  \factor[minimum size=0.3cm] {x} {} {} {};
  \node[latent, above=of x, minimum size=0.5cm] (z1)    {} ;
  \node (eps0) [minimum size=0.0cm,right=0.75cm of x] {};
  \node (eps0a) [minimum size=0.0cm,above=0.12cm of eps0.center] {};
  \node (eps0b) [minimum size=0.0cm,below=0.02cm of eps0.center, xshift=-0.2cm] {};
  \node (eps0c) [minimum size=0.0cm,below=0.02cm of eps0.center, xshift=0.2cm] {};
  \draw[] (eps0a.center)--(eps0b.center)--(eps0c.center)--(eps0a.center);
  \node[latent, left=0.5cm of z1, minimum size=0.5cm] (beta1)    {} ;

  \node[right=of x, xshift=-1.15cm, yshift=0.4cm] (xlabel) {$\mbx_n$};
  \node[right=0.05cm of z1] (z1label) {$\mbz_{n}$};
  \node[right=0.04cm of eps0] (eps0label) {$\mbepsilon_{n}$};
  \node[left=0.04cm of beta1] (beta1label) {$\beta$};

  \edge{z1}{x};
  \edge{eps0}{x};
  \edge{beta1}{z1};
  \edge{beta1}{x};

  \plate[inner sep=0.2cm,xshift=-0.2cm,
    label={[xshift=-15pt,yshift=15pt]south east:$N$}] {plate1} {
    (x)(xlabel)(z1)(z1label)(eps0)(eps0label)
  } {};

\end{tikzpicture}
  \label{sub:him}
\end{subfigure}%
\caption{(\textbf{left}) Hierarchical model, with local variables
$\mbz$ and global variables $\beta$.
(\textbf{right}) \textbf{Hierarchical implicit model}. It is a hierarchical
model where $\mbx$ is a deterministic function (denoted with a square) of noise $\mbepsilon$ (denoted with a triangle).}
\label{fig:hierarchical_model}
\vspace{-1em}
\end{figure}
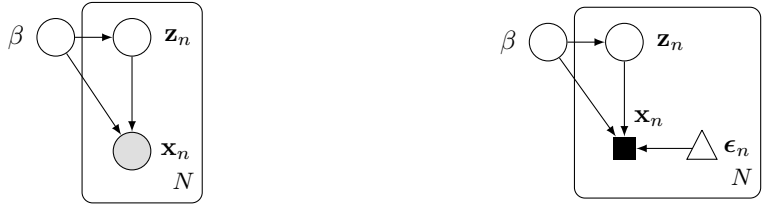

Hierarchical models play an
important role in sharing statistical strength across examples
\citep{gelman2006data}.  For a broad class of hierarchical Bayesian
models, the joint distribution of the hidden and
observed variables is
\begin{align}
p(\mbx, \mbz, \mbbeta) =
p(\mbbeta) \prod_{n=1}^N p(\mbx_n \g \mbz_n, \mbbeta) p(\mbz_n \g \mbbeta),
\label{eq:hm}
\end{align}
where $\mbx_n$ is an observation, $\mbz_n$ are latent variables
associated to that observation ({local variables}), and $\mbbeta$ are
latent variables shared across observations ({global variables}).
See \myfig{hierarchical_model} (left).

With hierarchical models,
local variables can be used for clustering in mixture
models, mixed memberships in topic models
\citep{blei2003latent}, and factors in probabilistic matrix
factorization \citep{salakhutdinov2008bayesian}.
Global variables can be used to pool information
across data points
for hierarchical regression \citep{gelman2006data},
topic models \citep{blei2003latent}, and
Bayesian nonparametrics \citep{teh2010hierarchical}.

Hierarchical models typically use a tractable likelihood
$p(\mbx_n \g \mbz_n, \mbbeta)$.
But many likelihoods of interest, such as
simulator-based models \citep{hartig2011statistical}
and \acrlongpl{GAN} \citep{goodfellow2014generative},
admit high fidelity to the true data
generating process and do not admit a tractable likelihood.
To overcome this limitation, we develop \glsreset{HIM}\emph{\glspl{HIM}}.

\Acrlongpl{HIM} have the same joint factorization as \myeqp{hm} but
only assume that one can sample from the likelihood.
Rather than define $p(\mbx_n \g \mbz_n, \mbbeta)$ explicitly,
\glspl{HIM} define a function $g$
that takes in random noise $\mbepsilon_n \sim s(\cdot)$
and outputs $\mbx_n$ given $\mbz_n$ and $\mbbeta$,
\begin{align*}
\mbx_n = g(\mbepsilon_n \g \mbz_n, \mbbeta),
\quad \mbepsilon_n\sim s(\cdot).
\end{align*}
The induced, implicit likelihood of $\mbx_n\in A$ given $\mbz_n$ and $\mbbeta$ is
\begin{align*}
\cP(\mbx_n \in A \g \mbz_n, \mbbeta) =
\int_{\{g(\mbepsilon_n\g \mbz_n, \mbbeta) =\mbx_n \in A\}}
s(\mbepsilon_n) \d\mbepsilon_n.
\end{align*}
This integral is typically intractable. It is difficult to find the
set to integrate over, and the integration itself may be expensive for
arbitrary noise distributions $s(\cdot)$ and functions $g$.

\myfig{hierarchical_model} (right) displays the graphical model for
\glspl{HIM}.  Noise ($\mbepsilon_n$) are denoted by triangles;
deterministic computation ($\mbx_n$) are denoted by squares.  We
illustrate two examples.

\parhead{Example: Physical Simulators.}  Given initial conditions,
simulators describe a stochastic process that generates data. For
example, in population ecology, the Lotka-Volterra model simulates
predator-prey populations over time via a stochastic differential
equation \citep{wilkinson2011stochastic}. For prey and predator
populations $x_1,x_2\in\mathbb{R}^+$ respectively, one process is
\begin{align*}
\frac{\d x_1}{\d t} &= \beta_1 x_1 - \beta_2 x_1 x_2 + \epsilon_1,
\quad
\hspace{0.8em}
\epsilon_1\sim \operatorname{Normal}(0, 10),
\\
\frac{\d x_2}{\d t} &= -\beta_2 x_2 + \beta_3 x_1 x_2 + \epsilon_2,
\quad
\epsilon_2\sim \operatorname{Normal}(0, 10),
\end{align*}
where Gaussian noises $\epsilon_1,\epsilon_2$ are added at each full time step.
The simulator runs for $T$ time steps given initial population sizes
for $x_1,x_2$.
Lognormal priors are placed over $\beta$.
The Lotka-Volterra model is grounded by theory but
features an intractable likelihood.
We study it in \Cref{sub:simulator}.

\paragraph{Example: Bayesian Generative Adversarial Network.}
\glsreset{GAN}\Glspl{GAN} define an implicit model and a method for
parameter estimation \citep{goodfellow2014generative}.  They are known
to perform well on image generation \citep{radford2016unsupervised}.
Formally, the implicit model for a \gls{GAN} is
\begin{align}
\mbx_n = g(\mbepsilon_n ; \mbtheta), \quad \mbepsilon_n \sim s(\cdot),
\label{eq:gan}
\end{align}
where $g$ is a neural network with parameters $\mbtheta$, and $s$ is a
standard normal or uniform. The neural network $g$ is typically not
invertible; this makes the likelihood intractable.

The parameters $\mbtheta$ in \glspl{GAN} are estimated by
divergence minimization between the generated and real data.
We make \glspl{GAN} amenable to Bayesian analysis
by placing a prior on the parameters $\mbtheta$.
We call this a Bayesian \gls{GAN}.
Bayesian \glspl{GAN}
enable modeling of parameter uncertainty and
are inspired by Bayesian neural networks, which have been shown to
improve the uncertainty and data efficiency of standard neural networks
\citep{mackay1992bayesian,neal1994bayesian}.
We study Bayesian \glspl{GAN} in \Cref{sub:gan};
Appendix B
provides example implementations in
the Edward probabilistic programming language
\citep{tran2016edward}.

\section{Likelihood-Free Variational Inference}
\label{sec:inference}

We described hierarchical implicit models, a rich class of latent
variable models with local and global structure alongside an
implicit density. Given data, we aim
to calculate the model's posterior
$p(\mbz, \mbbeta \g \mbx) = p(\mbx, \mbz, \mbbeta) / p(\mbx)$.
This is difficult as the normalizing constant $p(\mbx)$ is typically
intractable. With implicit models, the lack of a likelihood function
introduces an additional source of intractability.

We use variational inference \citep{jordan1999variational}.
It posits an approximating family $q\in\cQ$ and optimizes to find the
member closest to $p(\mbz,\mbbeta\g\mbx)$.
There are many choices of variational objectives
that measure closeness
\citep{ranganath2016operator,li2016variational,dieng2016chi}. To choose
an objective, we lay out desiderata for a variational
inference algorithm for implicit models:
\begin{enumerate}[leftmargin=*]
\item \emph{Scalability}.
Machine learning hinges on stochastic optimization to scale to
massive data \citep{bottou2010large}. The variational objective
should admit unbiased subsampling with the standard
technique,
\begin{equation*}
\sum_{n=1}^N f(\mbx_n) \approx \frac{N}{M} \sum_{m=1}^M f(\mbx_m),
\end{equation*}
where some computation $f(\cdot)$ over the full data is approximated
with a mini-batch of data $\{\mbx_m\}$.
\item \emph{Implicit Local Approximations}.
Implicit models specify flexible densities; this induces
very complex posterior distributions. Thus we would like a rich
approximating family for the per-data point approximations
$q(\mbz_n \g \mbx_n, \mbbeta)$. This means the variational objective
should only require that one can sample $\mbz_n\sim
q(\mbz_n\g\mbx_n,\mbbeta)$ and not evaluate its density.
\end{enumerate}
One variational objective meeting our desiderata is based on the
classical minimization of the \gls{KL} divergence.
(Surprisingly,
Appendix C
details how
the \gls{KL} is the
\emph{only} possible objective among a broad class.)

\subsection{KL Variational Objective}
\label{sub:kl}
Classical variational inference minimizes the \gls{KL} divergence from
the variational approximation $q$ to the posterior. This is equivalent
to maximizing the \gls{ELBO},
\begin{align}
\cL = \E_{q(\mbbeta,\mbz\g\mbx)}[\log p(\mbx, \mbz, \mbbeta) - \log q(\mbbeta, \mbz \g \mbx)].
\label{eq:elbo}
\end{align}
Let $q$ factorize in the same way as the posterior,
\begin{align*}
q(\mbbeta, \mbz \g \mbx) =
q(\mbbeta) \prod_{n=1}^N q(\mbz_n \g \mbx_n, \mbbeta),
\end{align*}
where $q(\mbz_n\g\mbx_n,\mbbeta)$ is an intractable density and
since the data $\mbx$ is constant during inference, we drop conditioning
for the global $q(\mbbeta)$.
Substituting $p$ and
$q$'s factorization yields
\begin{align*}
\cL &= \E_{q(\mbbeta)}[\log p(\mbbeta) - \log q(\mbbeta)] +
\sum_{n=1}^N \E_{q(\mbbeta) q(\mbz_n\g\mbx_n, \mbbeta)}
[\log p(\mbx_n,\mbz_n\g \mbbeta)- \log q(\mbz_n \g \mbx_n, \mbbeta)].
\end{align*}
This objective presents difficulties: the local densities
$p(\mbx_n,\mbz_n\g\mbbeta)$ and $q(\mbz_n \g \mbx_n, \mbbeta)$ are
both intractable.
To solve this, we consider ratio estimation.

\subsection{Ratio Estimation for the KL Objective}
\label{sub:ratio}
Let $q(\mbx_n)$ be the empirical distribution on the observations
$\mbx$ and consider using it in a ``variational joint''
$q(\mbx_n,\mbz_n\g\mbbeta)=q(\mbx_n)q(\mbz_n\g\mbx_n,\mbbeta)$.  Now
subtract the log empirical $\log q(\mbx_n)$ from the \gls{ELBO} above.
The \gls{ELBO} reduces to
\begin{align}
\begin{split}
\cL &\propto
\E_{q(\mbbeta)}[\log p(\mbbeta) - \log q(\mbbeta)] +
\sum_{n=1}^N
\E_{q(\mbbeta)q(\mbz_n\g\mbx_n,\mbbeta)}
\left[\log \frac{p(\mbx_n, \mbz_n\g
    \mbbeta)}{q(\mbx_n,\mbz_n\g\mbbeta)}\right].
\end{split}
\label{eq:joint-elbo}
\end{align}
(Here the proportionality symbol means equality up to additive
constants.) Thus the \gls{ELBO} is a function of the ratio of two
intractable densities. If we can form an estimator of this ratio, we
can proceed with optimizing the \gls{ELBO}.

We apply techniques for ratio estimation \citep{sugiyama2012density}.
It is a key idea in \glspl{GAN}
\citep{mohamed2016learning,uehara2016generative}, and
similar ideas have rearisen in statistics and physics
\citep{gutmann2014statistical,cranmer2015approximating}.
In particular, we use class probability estimation:
given a sample from $p(\cdot)$ or
$q(\cdot)$ we aim to estimate the probability that it belongs to
$p(\cdot)$. We model this using $\sigma(r(\cdot;\mbtheta))$, where $r$
is a parameterized function (e.g., neural network) taking sample
inputs and outputting a real value; $\sigma$ is the logistic
function outputting the probability.

We train $r(\cdot;\mbtheta)$ by minimizing a loss function
known as a proper scoring rule \citep{gneiting2007strictly}. For
example, in experiments we use the log loss,
\begin{equation}
\label{eq:log-loss}
\cD_{\textrm{log}} =
\mathbb{E}_{p(\mbx_n,\mbz_n\g\mbbeta)}
[- \log \sigma(r(\mbx_n, \mbz_n, \mbbeta; \mbtheta)) ] +
\mathbb{E}_{q(\mbx_n,\mbz_n\g\mbbeta)}
[- \log (1 - \sigma(r(\mbx_n, \mbz_n, \mbbeta; \mbtheta))) ].
\end{equation}
The loss is zero if $\sigma(r(\cdot;\mbtheta))$ returns 1 when
a sample is from $p(\cdot)$ and 0 when a sample is from $q(\cdot)$.
(We also experiment with the hinge loss; see \Cref{sub:simulator}.)
If $r(\cdot;\mbtheta)$ is sufficiently expressive, minimizing the
loss returns the optimal function \citep{mohamed2016learning},
\begin{equation*}
r^*(\mbx_n,\mbz_n,\mbbeta) =
\log p(\mbx_n,\mbz_n\g\mbbeta) - \log q(\mbx_n,\mbz_n\g\mbbeta).
\end{equation*}
As we minimize \myeqp{log-loss}, we use
$r(\cdot;\mbtheta)$ as a proxy to the log ratio in \myeqp{joint-elbo}.
Note $r$ estimates the log ratio; it's of direct interest and more numerically stable than the ratio.

The gradient of $\cD_{\textrm{log}}$ with respect to $\mbtheta$ is
\begin{align}
\begin{split}
\mathbb{E}_{p(\mbx_n,\mbz_n\g\mbbeta)}
[ \nabla_{\mbtheta} \log \sigma(r(\mbx_n, \mbz_n, \mbbeta; \mbtheta)) ] +
\mathbb{E}_{q(\mbx_n,\mbz_n\g\mbbeta)}
[ \nabla_{\mbtheta} \log (1 - \sigma(r(\mbx_n, \mbz_n, \mbbeta; \mbtheta))) ].
\end{split}
\label{eq:grad_d}
\end{align}
We compute unbiased gradients with Monte Carlo.

\subsection{Stochastic Gradients of the KL Objective}

To optimize the \gls{ELBO}, we use the ratio estimator,
\begin{align}
\begin{split}
\cL &=
\E_{q(\mbbeta\g\mbx)}[\log p(\mbbeta) - \log q(\mbbeta)] +
\sum_{n=1}^N
\E_{q(\mbbeta\g\mbx)q(\mbz_n\g\mbx_n,\mbbeta)}[ r(\mbx_n,\mbz_n,\beta) ].
\end{split}
\label{eq:ratio-elbo}
\end{align}
All terms are now tractable. We can calculate gradients to optimize
the variational family $q$.
Below we assume the priors $p(\mbbeta),p(\mbz_n\g\mbbeta)$ are
differentiable.
(We discuss methods to handle discrete global variables in the next section.)

We focus on reparameterizable variational approximations
\citep{kingma2014autoencoding,rezende2014stochastic}.
They enable sampling via a differentiable transformation $T$ of
random noise, $\delta\sim s(\cdot)$.
Due to \myeqp{ratio-elbo}, we require the global approximation
$q(\mbbeta;\mblambda)$ to admit a tractable density.
With reparameterization, its sample is
\begin{align*}
\mbbeta = \Tglobal(\deltaglobal ; \mblambda),
\quad \deltaglobal \sim \nglobal,
\end{align*}
for a choice of transformation $\Tglobal(\cdot;\mblambda)$ and noise
$\nglobal$.
For example, setting $\nglobal=\mathcal{N}(0,1)$ and
$\Tglobal(\deltaglobal)=\mu + \sigma\deltaglobal$ induces a normal
distribution $\mathcal{N}(\mu,\sigma^2)$.

Similarly for the local variables $\mbz_n$, we specify
\begin{align*}
\mbz_n = \Tlocal(\deltalocal, \mbx_n, \mbbeta; \mbphi), \quad \deltalocal \sim \nlocal.
\end{align*}
Unlike the global approximation, the local variational density
$q(\mbz_n\g\mbx_n;\mbphi)$ need not be tractable: the ratio estimator
relaxes this requirement.
It lets us leverage implicit
models not only for data but also for approximate posteriors.
In practice, we also amortize computation with inference networks,
sharing parameters $\phi$ across the per-data point approximate posteriors.

The gradient with respect to global parameters $\mblambda$ under this approximating family
is
\begin{align}
\begin{split}
\nabla_\mblambda \cL &=
\E_{s(\deltaglobal)}[\nabla_{\mblambda} (\log p(\mbbeta) - \log q(\mbbeta))]
] +
\sum_{n=1}^N
\E_{s(\deltaglobal)s_n(\deltalocal)}
[\nabla_{\mblambda} r(\mbx_n,\mbz_n,\beta) ].
\end{split}
\label{eq:grad_qglobal}
\end{align}
The gradient backpropagates through the local sampling
$\mbz_n = \Tlocal(\deltalocal, \mbx_n, \mbbeta; \mbphi)$ and the
global reparameterization $\mbbeta=\Tglobal(\deltaglobal;\mblambda)$.
We compute unbiased gradients with Monte Carlo.
The gradient with respect to local parameters $\mbphi$ is
\begin{align}
\label{eq:grad_qlocal}
\begin{split}
\nabla_\mbphi \cL &=
\sum_{n=1}^N \E_{q(\mbbeta) s(\deltalocal)}
[\nabla_{\mbphi} r(\mbx_n, \mbz_n, \mbbeta)].
\end{split}
\end{align}
where the gradient backpropagates through $\Tlocal$.%
\footnote{The ratio $r$ indirectly depends on $\mbphi$ but its gradient
w.r.t. $\mbphi$ disappears. This is derived via the score function
identity and the product rule (see, e.g.,
\citet[Appendix]{ranganath2014black}).}

\subsection{Algorithm}

\begin{algorithm}[t]
\SetKwInOut{Input}{Input}
\SetKwInOut{Output}{Output}
 \Input{Model $\mbx_n,\mbz_n\sim p(\cdot\g \mbbeta)$, $p(\mbbeta)$ \newline
Variational approximation
$\mbz_n\sim q(\cdot \g \mbx_n, \mbbeta ; \mbphi)$,
$q(\mbbeta \g \mbx ; \mblambda)$, \newline
Ratio estimator $r(\cdot;\mbtheta)$}
 \Output{Variational parameters $\mblambda$, $\mbphi$}
 Initialize $\mbtheta$, $\mblambda$, $\mbphi$ randomly.

 \While{not converged}{
  \vspace{0.25ex}
  Compute unbiased estimate of $\nabla_{\mbtheta} \cD$ (\myeqp{grad_d}), $\nabla_{\mblambda} \cL$ (\myeqp{grad_qglobal}), $\nabla_{\mbphi} \cL$ (\myeqp{grad_qlocal}). \\
  Update $\mbtheta$, $\mblambda$, $\mbphi$ using stochastic gradient descent. \\
 }
 \caption{\glsreset{LFVI}\Gls{LFVI}}
 \label{alg:vi}

\end{algorithm}

\myalg{vi} outlines the procedure. We call it \glsreset{LFVI}\emph{\gls{LFVI}}.
\gls{LFVI} is black box: it applies to models in which one can
simulate data and local variables, and calculate densities for the
global variables.
\gls{LFVI} first updates $\mbtheta$ to improve the ratio estimator $r$.
Then it uses $r$ to update parameters $\{\mblambda,\mbphi\}$ of the
variational approximation $q$. We optimize $r$ and $q$ simultaneously.
The algorithm is available in Edward
\citep{tran2016edward}.

\gls{LFVI} is scalable: we can
unbiasedly estimate the gradient over the
full data set with mini-batches \citep{hoffman2013stochastic}.
The algorithm can also handle models of either continuous or discrete data.
The requirement for differentiable global variables and reparameterizable
global approximations can be relaxed using score function gradients
\citep{ranganath2014black}.

Point estimates of the global parameters $\mbbeta$ suffice for many
applications \citep{goodfellow2014generative,rezende2014stochastic}.
\myalg{vi} can find point estimates: place a point
mass approximation $q$ on the parameters $\mbbeta$. This
simplifies gradients and corresponds to variational
EM.

\section{Experiments}
\label{sec:experiments}

We developed new models and inference.
For experiments, we study three applications: a large-scale
physical simulator for predator-prey populations in ecology; a
Bayesian \gls{GAN} for supervised classification; and a deep implicit
model for symbol generation.
In addition,
Appendix F,
provides practical advice on how
to address the stability of the ratio estimator by analyzing a toy
experiment.
We initialize parameters from a standard normal and apply gradient
descent with ADAM.

\parhead{Lotka-Volterra Predator-Prey Simulator.} We analyze the
Lotka-Volterra simulator of \Cref{sec:model} and follow the same setup
and hyperparameters of \citet{papamakarios2016fast}.
Its global variables $\mbbeta$ govern rates of change in a
simulation of predator-prey populations.
To infer them, we posit a mean-field normal approximation
(reparameterized to be on the same support) and run \myalg{vi} with
both a log loss and hinge loss for the ratio estimation problem;
Appendix D
details the hinge loss.
We compare to rejection ABC, MCMC-ABC, and SMC-ABC
\citep{marin2012approximate}.
MCMC-ABC uses a spherical Gaussian proposal; SMC-ABC is manually tuned with a decaying epsilon schedule; all ABC methods are tuned to use the best performing hyperparameters such as the tolerance error.

\label{sub:simulator}

\begin{figure}[t]
\begin{subfigure}{0.5\columnwidth}
  \centering
  \includegraphics[width=0.7\columnwidth]{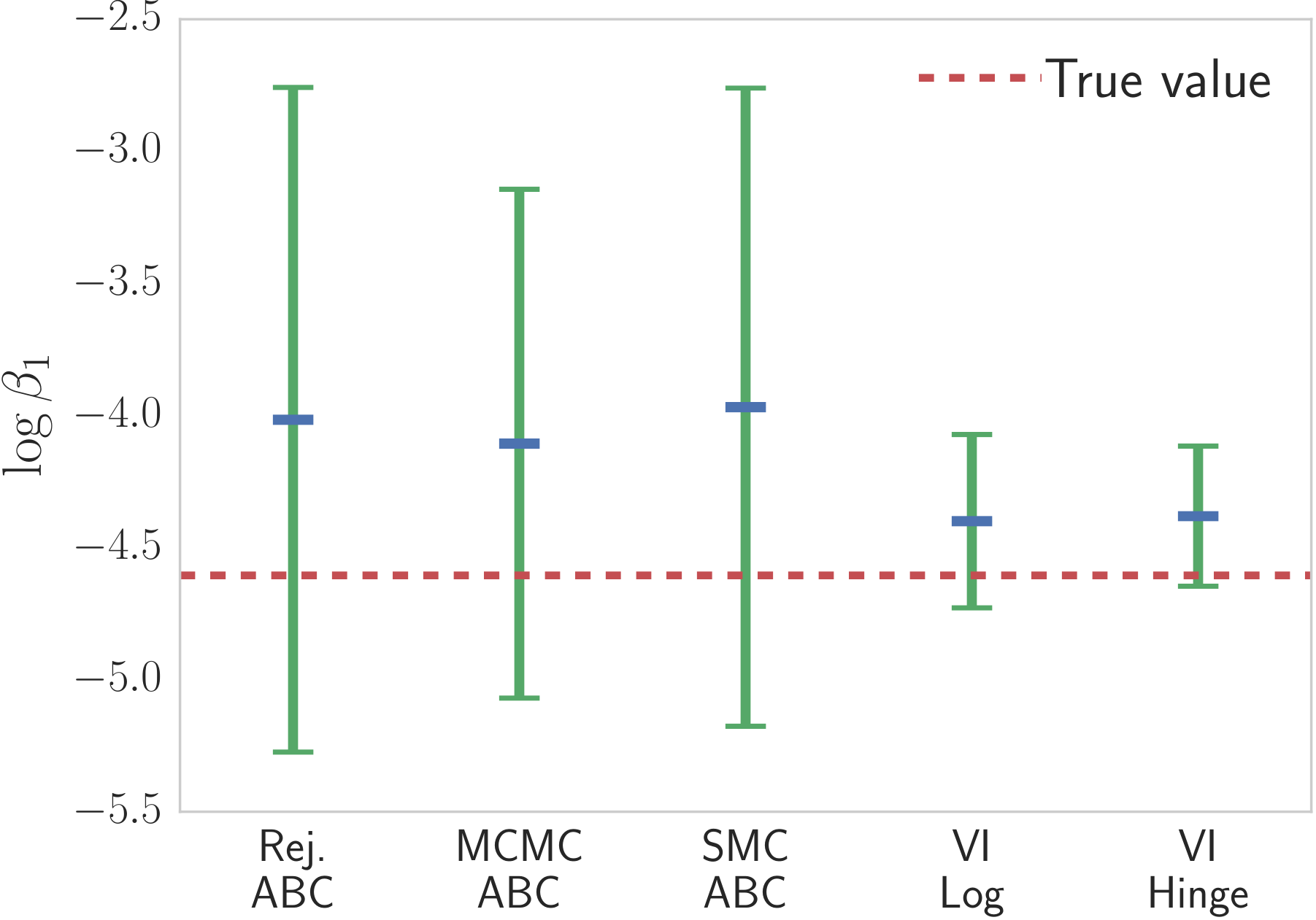}
\end{subfigure}%
\begin{subfigure}{0.5\columnwidth}
  \centering
  \includegraphics[width=0.7\columnwidth]{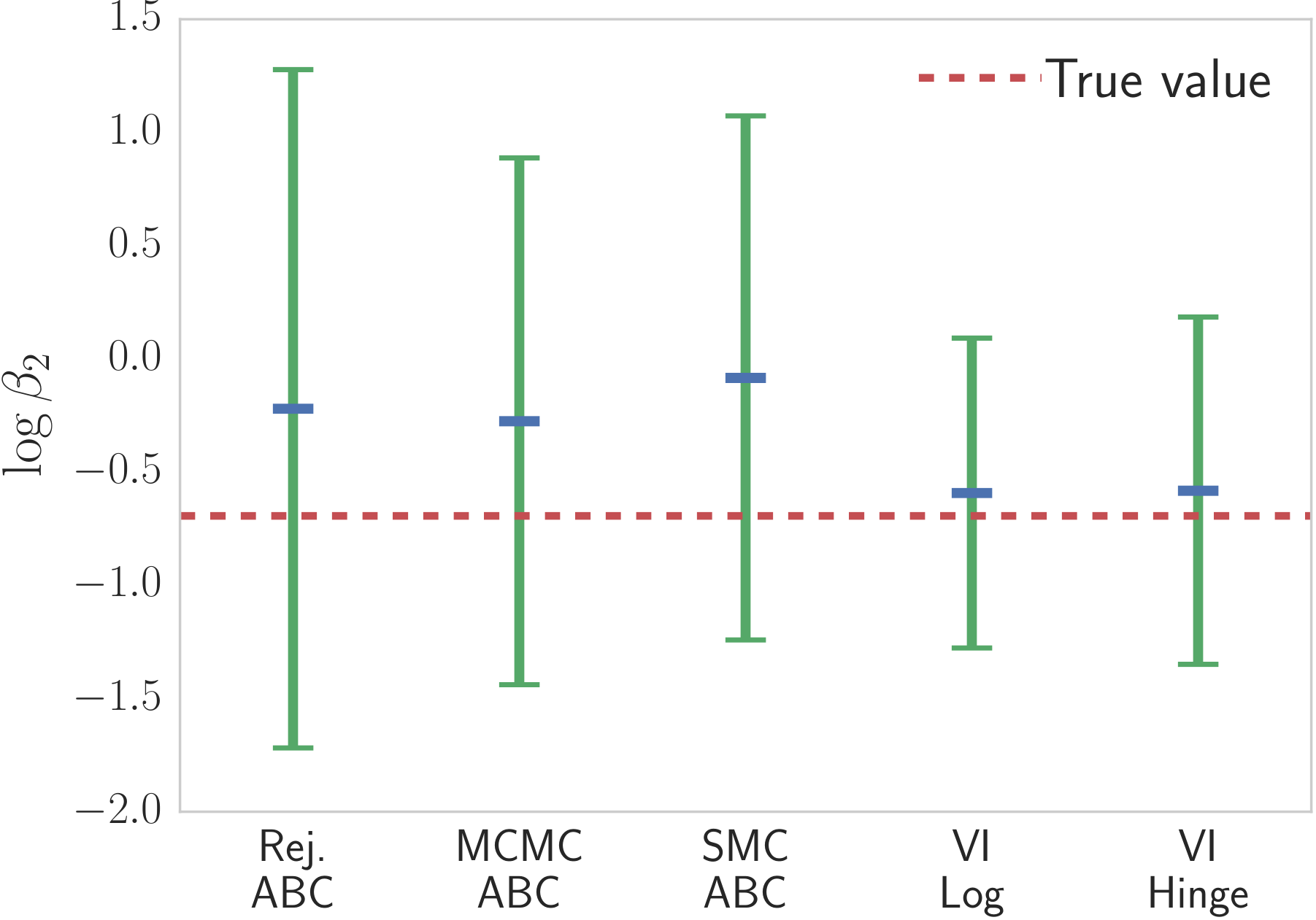}
\end{subfigure}
\begin{subfigure}{0.5\columnwidth}
  \centering
  \includegraphics[width=0.7\columnwidth]{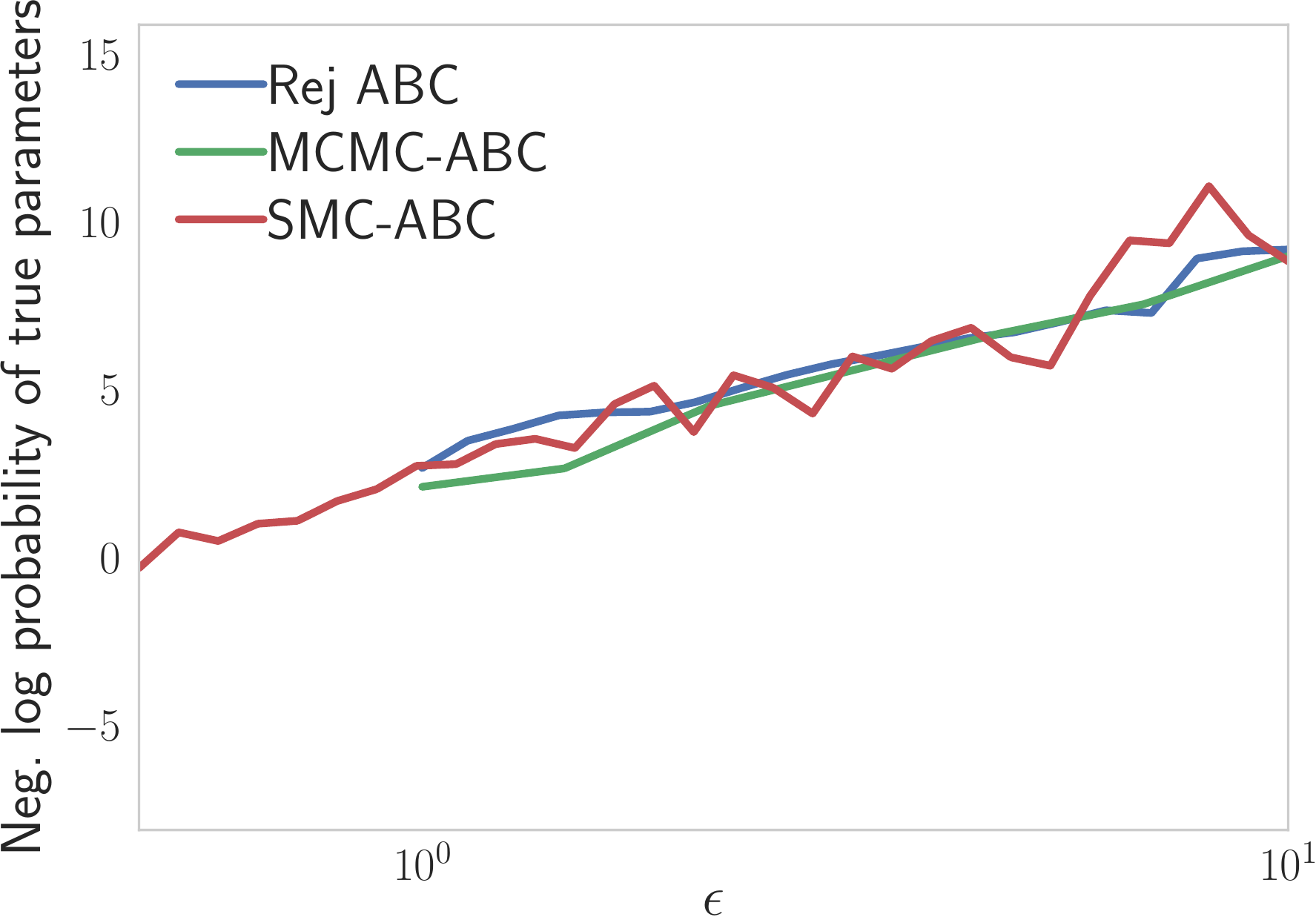}
\end{subfigure}%
\begin{subfigure}{0.5\columnwidth}
  \centering
  \includegraphics[width=0.7\columnwidth]{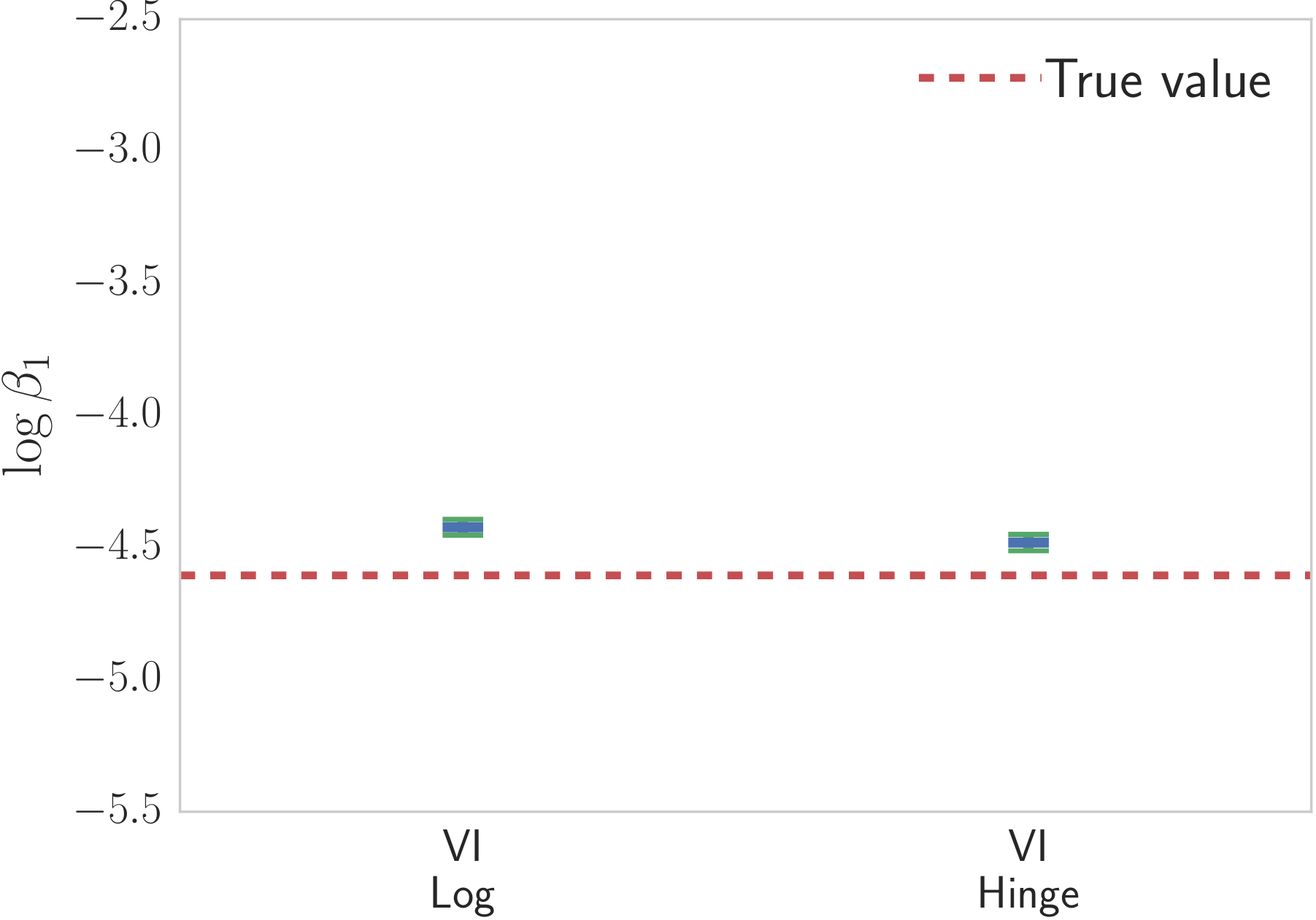}
\end{subfigure}
\caption{\textbf{(top)}
Marginal posterior for first two parameters.
\textbf{(bot. left)}
ABC methods over tolerance error.
\textbf{(bot. right)}
Marginal posterior for first parameter on a large-scale data set.
Our inference achieves more accurate results and scales to massive
data.
}
\label{fig:simulator}
\vspace{-1.25ex}
\end{figure}

\myfig{simulator} displays results on two data sets.
In the top figures and bottom left, we analyze data consisting of a
simulation for $T=30$ time steps, with recorded values of the
populations every $0.2$ time units.
The bottom left figure calculates the negative log probability of the
true parameters over the tolerance error for \gls{ABC} methods;
smaller tolerances result in more accuracy but
slower runtime. The top figures compare the marginal posteriors for two
parameters using the smallest tolerance for the \gls{ABC} methods.
Rejection ABC, MCMC-ABC, and SMC-ABC all contain the true parameters
in their 95\% credible interval but are less
confident than our methods. Further, they required $100,000$
simulations from the model, with an acceptance rate of $0.004\%$ and
$2.990\%$ for rejection ABC and MCMC-ABC respectively.

The bottom right figure analyzes data consisting of $100,000$
time series, each of the same size as the single time series analyzed
in the previous figures. This size is not possible with traditional methods.
Further, we see that with our methods, the posterior concentrates
near the truth. We also experienced little difference in accuracy
between using the log loss or the hinge loss for ratio estimation.

\begin{table*}[tb]
\centering
\begin{tabular}{llllll}
\toprule
& \multicolumn{4}{c}{Test Set Error} \\
Model + Inference & Crabs & Pima & Covertype & MNIST \\
\midrule
Bayesian GAN + VI & 0.03 & \textbf{0.232} & \textbf{0.154} & \textbf{0.0136} \\
Bayesian GAN + MAP & 0.12 & 0.240 & 0.185 & 0.0283 \\
Bayesian NN + VI & \textbf{0.02} & 0.242 & 0.164 & 0.0311 \\
Bayesian NN + MAP & 0.05 & 0.320 & 0.188 & 0.0623 \\
\bottomrule
\end{tabular}
\caption{Classification accuracy of Bayesian \gls{GAN} and Bayesian neural networks
across small to medium-size data sets. Bayesian \glspl{GAN}
achieve comparable or better performance to their Bayesian neural net
counterpart.}
\label{table:classification}
\vspace{-1em}
\end{table*}

\label{sub:gan}

\parhead{Bayesian Generative Adversarial Networks.} We analyze
Bayesian \glspl{GAN}, described in \Cref{sec:model}.
Mimicking a use case of Bayesian neural networks
\citep{blundell2015weight,hernandez-lobato2016black}, we apply
Bayesian \glspl{GAN} for classification on small to medium-size data.
The \gls{GAN} defines a conditional $p(y_n\g\mbx_n)$,
taking a feature $\mbx_n\in\mathbb{R}^D$ as input
and generating a label $y_n\in\{1,\ldots,K\}$, via the process
\begin{equation}
\label{eq:bayesian-gan-exp}
y_n = g(\mbx_n,\mbepsilon_n\g\mbtheta),
\qquad
\mbepsilon_n\sim\mathcal{N}(0,1),
\end{equation}
where $g(\cdot\g\mbtheta)$ is a 2-layer multilayer perception
with ReLU activations, batch normalization, and is
parameterized by weights and biases $\mbtheta$.
We place normal priors,
$\mbtheta\sim\mathcal{N}(0,1)$.

We analyze two choices of the variational model: one with a mean-field
normal approximation for
$q(\mbtheta\g\mbx)$, and another with a point mass approximation
(equivalent to maximum a posteriori). We
compare to a Bayesian neural network, which uses the same generative
process as \myeqp{bayesian-gan-exp} but draws from a Categorical
distribution rather than feeding noise into the neural net.  We fit it
separately using a mean-field normal approximation and
maximum a posteriori. \mytable{classification} shows that Bayesian
\glspl{GAN} generally outperform their Bayesian neural net
counterpart.

Note that
Bayesian \glspl{GAN} can analyze discrete data such as in generating a
classification label.
Traditional \glspl{GAN} for discrete data is an open challenge
\citep{kusner2016gans}.
In
Appendix E,
we compare Bayesian
\glspl{GAN} with point estimation to typical \glspl{GAN}.
Bayesian \glspl{GAN} are also able to leverage parameter
uncertainty for analyzing these small to medium-size data sets.

One problem with Bayesian \glspl{GAN} is that
they cannot work with very large neural networks: the ratio estimator
is a function of global parameters, and thus the input size grows with
the size of the neural network. One approach is to make
the ratio estimator not a function of the global parameters. Instead
of optimizing model parameters via variational EM, we can train the
model parameters by backpropagating through the ratio objective
instead of the variational objective.
An alternative is to use the
hidden units as input which is much lower
dimensional \citep[Appendix~C]{tran2017implicit}.

\parhead{Injecting Noise into Hidden Units.} In this section, we show how to build a hierarchical implicit model by
simply injecting randomness into hidden units.
We model sequences $\mbx=(\mbx_1,\ldots,\mbx_T)$ with a recurrent
neural network.  For $t=1,\ldots,T$,
\begin{align*}
\mbz_{t} &= g_z(\mbx_{t-1}, \mbz_{t-1}, \mbepsilon_{t,z}),
\quad
\mbepsilon_{t,z}\sim\mathcal{N}(0,1),
\\
\mbx_{t} &= g_x(\mbz_{t}, \mbepsilon_{t,x}),
\quad
\hspace{3.35em}
\mbepsilon_{t,x}\sim\mathcal{N}(0,1),
\end{align*}
where $g_z$ and $g_x$
are both 1-layer multilayer perceptions with ReLU activation and layer normalization.
We place standard normal priors over all weights and biases.
See \myfig{sequence-model}.

\begin{figure}[t]
\begin{subfigure}{0.5\columnwidth}
  \centering
  \begin{tikzpicture}

  \factor[minimum size=0.3cm] {xprev} {} {} {};
  \factor[right=1.2cm of xprev, minimum size=0.3cm] {xt} {} {} {};
  \factor[right=1.2cm of xt, minimum size=0.3cm] {xnext} {} {} {};

  \factor[above=1.2cm of x, minimum size=0.3cm] {zprev} {} {} {};
  \factor[right=1.2cm of zprev, minimum size=0.3cm] {zt} {} {} {};
  \factor[right=1.2cm of zt, minimum size=0.3cm] {znext} {} {} {};

  \node[left=of zprev, minimum size=0.5cm] (zprevdots) {$\cdots$};
  \node[right=of znext, minimum size=0.5cm] (znextdots) {$\cdots$};

  \node (epsprev) [minimum size=0.0cm,above=0.25cm of zprev, xshift=-0.5cm] {};
  \node (epspreva) [minimum size=0.0cm,above=0.12cm of epsprev.center] {};
  \node (epsprevb) [minimum size=0.0cm,below=0.02cm of epsprev.center, xshift=-0.2cm] {};
  \node (epsprevc) [minimum size=0.0cm,below=0.02cm of epsprev.center, xshift=0.2cm] {};
  \draw[] (epspreva.center)--(epsprevb.center)--(epsprevc.center)--(epspreva.center);

  \node (epst) [minimum size=0.0cm,above=0.25cm of zt, xshift=-0.5cm] {};
  \node (epsta) [minimum size=0.0cm,above=0.12cm of epst.center] {};
  \node (epstb) [minimum size=0.0cm,below=0.02cm of epst.center, xshift=-0.2cm] {};
  \node (epstc) [minimum size=0.0cm,below=0.02cm of epst.center, xshift=0.2cm] {};
  \draw[] (epsta.center)--(epstb.center)--(epstc.center)--(epsta.center);

  \node (epsnext) [minimum size=0.0cm,above=0.25cm of znext, xshift=-0.5cm] {};
  \node (epsnexta) [minimum size=0.0cm,above=0.12cm of epsnext.center] {};
  \node (epsnextb) [minimum size=0.0cm,below=0.02cm of epsnext.center, xshift=-0.2cm] {};
  \node (epsnextc) [minimum size=0.0cm,below=0.02cm of epsnext.center, xshift=0.2cm] {};
  \draw[] (epsnexta.center)--(epsnextb.center)--(epsnextc.center)--(epsnexta.center);

  \node (epsxprev) [minimum size=0.0cm,above=0.25cm of xprev, xshift=-0.5cm] {};
  \node (epsxpreva) [minimum size=0.0cm,above=0.12cm of epsxprev.center] {};
  \node (epsxprevb) [minimum size=0.0cm,below=0.02cm of epsxprev.center, xshift=-0.2cm] {};
  \node (epsxprevc) [minimum size=0.0cm,below=0.02cm of epsxprev.center, xshift=0.2cm] {};
  \draw[] (epsxpreva.center)--(epsxprevb.center)--(epsxprevc.center)--(epsxpreva.center);

  \node (epsxt) [minimum size=0.0cm,above=0.25cm of xt, xshift=-0.5cm] {};
  \node (epsxta) [minimum size=0.0cm,above=0.12cm of epsxt.center] {};
  \node (epsxtb) [minimum size=0.0cm,below=0.02cm of epsxt.center, xshift=-0.2cm] {};
  \node (epsxtc) [minimum size=0.0cm,below=0.02cm of epsxt.center, xshift=0.2cm] {};
  \draw[] (epsxta.center)--(epsxtb.center)--(epsxtc.center)--(epsxta.center);

  \node (epsxnext) [minimum size=0.0cm,above=0.25cm of xnext, xshift=-0.5cm] {};
  \node (epsxnexta) [minimum size=0.0cm,above=0.12cm of epsxnext.center] {};
  \node (epsxnextb) [minimum size=0.0cm,below=0.02cm of epsxnext.center, xshift=-0.2cm] {};
  \node (epsxnextc) [minimum size=0.0cm,below=0.02cm of epsxnext.center, xshift=0.2cm] {};
  \draw[] (epsxnexta.center)--(epsxnextb.center)--(epsxnextc.center)--(epsxnexta.center);

  \node[right=0.05cm of xprev] (xprevlabel) {$\mbx_{t-1}$};
  \node[right=0.05cm of xt] (xtlabel) {$\mbx_{t}$};
  \node[right=0.05cm of xnext] (xnextlabel) {$\mbx_{t+1}$};

  \node[below=0.10cm of zprev, right=0.10cm] (zprevlabel) {$\mbz_{t-1}$};
  \node[below=0.10cm of zt, right=0.10cm] (ztlabel) {$\mbz_{t}$};
  \node[below=0.10cm of znext, right=0.10cm] (zTlabel) {$\mbz_{t+1}$};

  \edge{zprev}{xprev};
  \edge{zt}{xt};
  \edge{znext}{xnext};

  \edge{xprev}{zt};
  \edge{xt}{znext};

  \edge{zprevdots}{zprev};
  \edge{zprev}{zt};
  \edge{zt}{znext};
  \edge{znext}{znextdots};

  \edge{epsprev}{zprev};
  \edge{epsxprev}{xprev};
  \edge{epst}{zt};
  \edge{epsxt}{xt};
  \edge{epsnext}{znext};
  \edge{epsxnext}{xnext};

\end{tikzpicture}
  \vspace{6.5ex}
  \caption{A deep implicit model for sequences. It is a
  \gls{RNN} with noise injected into each hidden state.
  The hidden state is now an implicit latent variable.
  The same occurs for generating outputs.
  }
  \label{fig:sequence-model}
\end{subfigure}%
\begin{subfigure}{0.5\columnwidth}
  \begin{lstlisting}[basicstyle=\Large]
  -x+x/x**x*//x*x+
  x/x*x+x*x/x+x+x+
  /+x*x+x*x/x/x+x+
  /x+*x+x*x/x+x-x+
  x/x*x/x*x+x+x+x-
  x+x+x/x*x*x+x/x+\end{lstlisting}
  \vspace{-1ex}
  \caption{
  Generated symbols from the implicit model. Good samples
  place arithmetic operators between the variable $x$.
  The implicit model learned to follow rules from the context free grammar
  up to some multiple operator repeats.
  }
  \label{fig:sequence}
\end{subfigure}
\vspace{-1em}
\end{figure}

If the injected noise $\mbepsilon_{t,z}$ combines linearly with the
output of $g_z$, the induced distribution
$p(\mbz_t\g\mbx_{t-1},\mbz_{t-1})$ is Gaussian parameterized by that
output.
This defines a stochastic \gls{RNN}
\citep{bayer2014learning,fraccaro2016sequential}, which generalizes
its deterministic connection.
With nonlinear combinations, the implicit density is more flexible (and
intractable), making previous methods for inference not applicable.
In our method, we perform variational inference and specify $q$ to be
implicit; we use the same architecture as the probability model's
implicit priors.

We follow the same setup and hyperparameters as \citet{kusner2016gans}
and generate simple one-variable arithmetic sequences
following a context free grammar,
\begin{equation*}
S \to x \| S + S \| S - S \| S * S \| S/S,
\end{equation*}
where $\|$ divides possible productions of the grammar.
We concatenate the inputs and
point estimate the global variables (model parameters) using variational EM.
\myfig{sequence} displays samples from the inferred model, training on
sequences with a maximum of 15 symbols. It achieves sequences which
roughly follow the context free grammar.

\vspace{-1ex}
\section{Discussion}
\vspace{-1.0ex}
We developed a class of hierarchical implicit models and
likelihood-free variational inference, merging the idea of implicit
densities with hierarchical Bayesian modeling and approximate
posterior inference.  This expands Bayesian analysis with the ability
to apply neural samplers, physical simulators, and their combination
with rich, interpretable latent structure.

More stable inference with ratio estimation is an open challenge.
This is especially important when we
analyze large-scale real world
applications of implicit models. Recent work for genomics offers
a promising solution \citep{tran2017implicit}.

\paragraph{Acknowledgements.}
We thank Balaji Lakshminarayanan for discussions which helped motivate
this work. We also thank Christian Naesseth, Jaan Altosaar, and Adji
Dieng for their feedback and comments.
DT is supported by a Google Ph.D. Fellowship in Machine Learning and
an Adobe Research Fellowship.
This work is also supported by NSF IIS-0745520, IIS-1247664, IIS-1009542, ONR N00014-11-1-0651, DARPA FA8750-14-2-0009, N66001-15-C-4032, Facebook, Adobe, Amazon, and the John Templeton Foundation.

\bibliographystyle{apalike}
\bibliography{bib}

\appendix
\section{Noise versus Latent Variables}
\label{appendix:noise}

\Glspl{HIM} have two sources of randomness for each data point:
the latent variable $\mbz_n$ and the noise $\mbepsilon_n$; these
sources of randomness get transformed to produce $\mbx_n$.
Bayesian analysis infers posteriors on latent variables.
A natural question is whether one should
also infer the posterior of the noise.

The posterior's shape---and ultimately if it is
meaningful---is determined by the
dimensionality of noise and the transformation.
For example, consider the \gls{GAN} model, which has
no local latent variable, $\mbx_n=g(\mbepsilon_n ; \mbtheta)$.
The conditional $p(\mbx_n\g\mbepsilon_n)$ is a point mass, fully
determined by $\mbepsilon_n$.
When $g(\cdot;\mbtheta)$ is injective, the posterior
$p(\mbepsilon_n\g\mbx_n)$ is also a point mass,
\begin{align*}
p(\mbepsilon_n \g \mbx_n) &= \mathbb{I}[\mbepsilon_n = g^{-1}(\mbx_n)],
\end{align*}
where $g^{-1}$ is the left inverse of $g$.This means for injective functions of the randomness (both noise and latent variables), the ``posterior'' may be worth
analysis as a deterministic hidden representation
\citep{donahue2017adversarial}, but it is not random.

The point mass posterior can be found via
nonlinear least squares.
Nonlinear least squares yields the iterative algorithm
\begin{align*}
\hat{\mbepsilon}_n = \hat{\mbepsilon}_n - \rho_t \nabla_{\hat{\mbepsilon}_n}
f(\hat{\mbepsilon}_n)^\top (f(\hat{\mbepsilon}_n) - \mbx_n),
\end{align*}
for some step size sequence $\rho_t$. Note the updates will get stuck when
the gradient of $f$ is zero. However, the injective property of $f$ allows
the iteration to be checked for correctness (simply check if
$f(\hat{\mbepsilon}_n)=\mbx_n$).

\section{Implicit Model Examples in Edward}
\label{appendix:dim}

We demonstrate implicit models via example
implementations in Edward \citep{tran2016edward}.

\myfig{dim-edward} implements a 2-layer \acrlong{DIM}.
It uses \texttt{tf.layers} to define neural networks:
\texttt{tf.layers.dense(x, 256)}
applies a fully connected layer with $256$ hidden units and input $x$;
weight and bias parameters are abstracted from the user.
The program generates $N$ data points $\mbx_n\in\mathbb{R}^{10}$
using two layers of implicit latent variables
$\mbz_{n,1},\mbz_{n,2}\in\mathbb{R}^d$ and with an implicit
likelihood.

\myfig{bayesian-gan-edward} implements a Bayesian \gls{GAN} for
classification.  It manually defines a 2-layer neural network, where
for each data index, it takes features $\mbx_n\in\mathbb{R}^{500}$
concatenated with noise $\mbepsilon_n\in\mathbb{R}$ as input.
The output is a label $\mby_n\in\{-1,1\}$, given by
the sign of the last layer.
We place a standard normal prior over all weights and biases.
Running this program while feeding the placeholder
$\mbX\in\mathbb{R}^{N\times 500}$ generates a vector of labels
$\mby\in\{-1,1\}^N$.

\begin{figure}[!htb]
\begin{lstlisting}[language=python]
import tensorflow as tf
from edward.models import Normal

# random noise is Normal(0, 1)
eps2 = Normal(tf.zeros([N, d]), tf.ones([N, d]))
eps1 = Normal(tf.zeros([N, d]), tf.ones([N, d]))
eps0 = Normal(tf.zeros([N, d]), tf.ones([N, d]))

# alternate latent layers z with hidden layers h
z2 = tf.layers.dense(eps2, 128, activation=tf.nn.relu)
h2 = tf.layers.dense(z2, 128, activation=tf.nn.relu)
z1 = tf.layers.dense(tf.concat([eps1, h2], 1), 128, activation=tf.nn.relu)
h1 = tf.layers.dense(z1, 128, activation=tf.nn.relu)
x  = tf.layers.dense(tf.concat([eps0, h1], 1), 10, activation=None)
\end{lstlisting}
\caption{Two-layer \acrlong{DIM} for data points
$\mbx_n\in\mathbb{R}^{10}$. The architecture alternates with
stochastic and deterministic layers. To define a stochastic layer, we
simply inject noise by concatenating it into the input of a neural net
layer.}
\label{fig:dim-edward}
\end{figure}

\begin{figure}[!htb]
\begin{lstlisting}[language=python]
import tensorflow as tf
from edward.models import Normal

# weights and biases have Normal(0, 1) prior
W1 = Normal(tf.zeros([501, 256]), tf.ones([501, 256]))
W2 = Normal(tf.zeros([256, 1]), tf.ones([256, 1]))
b1 = Normal(tf.zeros(256), tf.ones(256))
b2 = Normal(tf.zeros(1), tf.ones(1))

# set up inputs to neural network
X = tf.placeholder(tf.float32, [N, 500])
eps = Normal(tf.zeros([N, 1]), tf.ones([N, 1]))

# y = neural_network([x, eps])
input = tf.concat([X, eps], 1)
h1 = tf.nn.relu(tf.matmul(input, W1) + b1)
h2 = tf.matmul(h1, W2) + b2
y = tf.reshape(tf.sign(h2), [-1])  # take sign, then flatten
\end{lstlisting}
\caption{Bayesian \gls{GAN} for classification, taking
$\mbX\in\mathbb{R}^{N\times 500}$ as input and generating a vector of
labels $\mby\in\{-1,1\}^N$. The neural network directly generates the data
rather than parameterizing a probability distribution.}
\label{fig:bayesian-gan-edward}
\end{figure}

\section{KL Uniqueness}
\label{appendix:uniqueness}

An integral probability metric measures distance between two
distributions $p$ and $q$,
\begin{align*}
d(p, q) = \sup_{f \in \cF} |\E_p f - \E_q f|.
\end{align*}
Integral probability metrics have been used for parameter estimation
in generative models \citep{dziugaite2015training} and for variational inference
in models with tractable density \citep{ranganath2016hierarchical}. In
contrast to models with only local latent variables,
to infer the posterior, we need an integral probability metric between
it and the variational approximation. The direct approach fails because
sampling from the posterior is intractable.

An indirect approach requires constructing a sufficiently broad class
of functions with posterior expectation zero based on Stein's method
\citep{ranganath2016hierarchical}.  These constructions require a
likelihood function and its gradient. Working around the likelihood
would require a form of nonparametric density estimation; unlike ratio
estimation, we are unaware of a solution that sufficiently scales to
high dimensions.

As another class of divergences, the $f$ divergence is
\begin{align*}
d(p, q) = \E_q\Big[f\left(\frac{p}{q}\right)\Big].
\end{align*}
Unlike integral probability metrics, $f$ divergences are naturally
conducive to ratio estimation, enabling implicit $p$ and implicit $q$.
However, the challenge lies in scalable computation. To subsample
data in hierarchical models, we need $f$ to satisfy up to constants
$f(ab) = f(a) +
f(b)$, so that the expectation becomes a sum over individual data
points. For
continuous functions, this is a defining property of the $\log$
function. This implies the KL-divergence from $q$ to $p$ is the only
$f$ divergence where the subsampling technique in our desiderata
is possible.

\section{Hinge Loss}
\label{appendix:hinge}

Let $r(\mbx_i,\mbz_i,\mbbeta;\theta)$ output a real value, as with the
log loss in
Section 4.
The hinge loss is
\begin{align*}
\cD_{\textrm{hinge}} &=
\mathbb{E}_{p(\mbx_n,\mbz_n\g\mbbeta)}
[ \max(0, 1 - r(\mbx_n, \mbz_n, \mbbeta; \mbtheta)) ] +
\\
&\hspace{1.35em}
\mathbb{E}_{q(\mbx_n,\mbz_n\g\mbbeta)}
[ \max(0, 1 + r(\mbx_n, \mbz_n, \mbbeta; \mbtheta)) ].
\end{align*}
We minimize this loss function by following unbiased gradients.
The gradients are calculated analogously as for the log loss. The optimal
$r^*$ is the log ratio.

\section{Comparing Bayesian GANs with MAP to GANs with MLE}
\label{appendix:comparing}

In Section 4,
we argued that MAP estimation with a Bayesian
\gls{GAN} enables analysis over discrete data, but \glspl{GAN}---even with a
maximum likelihood objective \citep{goodfellow2014on}---cannot.  This
is a surprising result: assuming a flat prior for MAP, the two
are ultimately optimizing the same objective.
We compare the two below.

For \glspl{GAN},
assume the discriminator outputs a logit probability, so that it's
unconstrained instead of on $[0,1]$.
\glspl{GAN} with MLE use the discriminative problem
\begin{equation*}
\max_\mbtheta
\mathbb{E}_{q(\mbx)} [ \log \sigma(D(\mbx; \mbtheta)) ] +
\mathbb{E}_{p(\mbx; \mbw)} [ \log (1 - \sigma(D(\mbx; \mbtheta))) ].
\end{equation*}
They use the generative problem
\begin{equation*}
\min_\mbw
\mathbb{E}_{p(\mbx;\mbw)} [ -\exp( D (\mbx) ) ].
\end{equation*}
Solving the generative problem with reparameterization gradients
requires backpropagating through data generated from the model,
$\mbx\sim p(\mbx;\mbw)$.
This is not possible for discrete $\mbx$. Further, the
exponentiation
also makes this objective numerically unstable and thus unusable in
practice.

Contrast this with Bayesian \glspl{GAN} with MLE (MAP and a flat prior).
This applies a point mass variational approximation
$q(\mbw') = \mathbb{I}[\mbw' = \mbw]$.
It maximizes the \gls{ELBO},
\begin{equation*}
\max_\mbw
\mathbb{E}_{q(\mbw)} [ \log p(\mbw) - \log q(\mbw) ] +
\sum_{n=1}^N r (\mbx_n, \mbw).
\end{equation*}
The first term is zero for a flat prior $p(\mbw)\propto 1$ and point
mass approximation; the problem reduces to
\begin{equation*}
\max_\mbw
\sum_{n=1}^N r (\mbx_n, \mbw).
\end{equation*}
Solving this is possible for discrete $\mbx$: it only requires
backpropagating gradients through $r(\mbx, \mbw)$ with respect to
$\mbw$, all of which is differentiable. Further, the objective does
not require a numerically unstable exponentiation.

Ultimately, the difference lies in the role of the ratio estimators.
Recall for Bayesian \glspl{GAN}, we use the ratio estimation problem
\begin{align*}
\mathcal{D}_{\rm log}
&
=
\mathbb{E}_{p(\mbx ; \mbw)} [ -\log \sigma(r(\mbx, \mbw; \mbtheta)) ] + \\
&
\mathbb{E}_{q(\mbx)} [ -\log (1 - \sigma(r(\mbx,\mbw; \mbtheta))) ].
\end{align*}
The optimal ratio estimator is the log-ratio
$r^*(\mbx,\mbw) = \log p(\mbx\g\mbw) - \log q(\mbx)$.
Optimizing it with respect to $\mbw$ reduces to optimizing
the log-likelihood $\log p(\mbx\g\mbw)$.
The optimal discriminator for
\glspl{GAN} with MLE has the same ratio,
$D^*(\mbx) = \log p(\mbx;\mbw) - \log q(\mbx)$; however, it is a constant
function with respect to $\mbw$. Hence one cannot immediately
substitute $D^*(\mbx)$ as a proxy to optimizing the likelihood. An
alternative is to
use importance sampling; the result is the former objective
\citep{goodfellow2014on}.

\section{Stability of Ratio Estimator}
\label{appendix:stability}

With implicit models, the difference from standard KL variational
inference lies in the ratio estimation problem. Thus we would like to assess the
accuracy of the ratio estimator. We can check this by comparing to the true
ratio under a model with tractable likelihood.

We apply Bayesian linear regression. It features a tractable posterior which
we leverage in our analysis.
We use 50 simulated data points
$\{\mby_n\in\mathbb{R}^2,\mbx_n\in\mathbb{R}\}$.
The optimal (log) ratio is
\begin{equation*}
r^*(\mbx,\mbbeta) =
\log p(\mbx\g\mbbeta) - \log q(\mbx).
\end{equation*}
Note the log-likelihood $\log p(\mbx\g\mbbeta)$ minus
$r^*(\mbx,\mbbeta)$ is equal to
the empirical distribution $\sum_n \log q(\mbx_n)$, a constant.
Therefore if a ratio estimator $r$ is accurate, its difference with
$\log p(\mbx\g\mbbeta)$ should be a
constant with low variance across values of $\mbbeta$.

\begin{figure}[t]
\begin{subfigure}{0.33\columnwidth}
  \centering
  \includegraphics[width=\columnwidth]{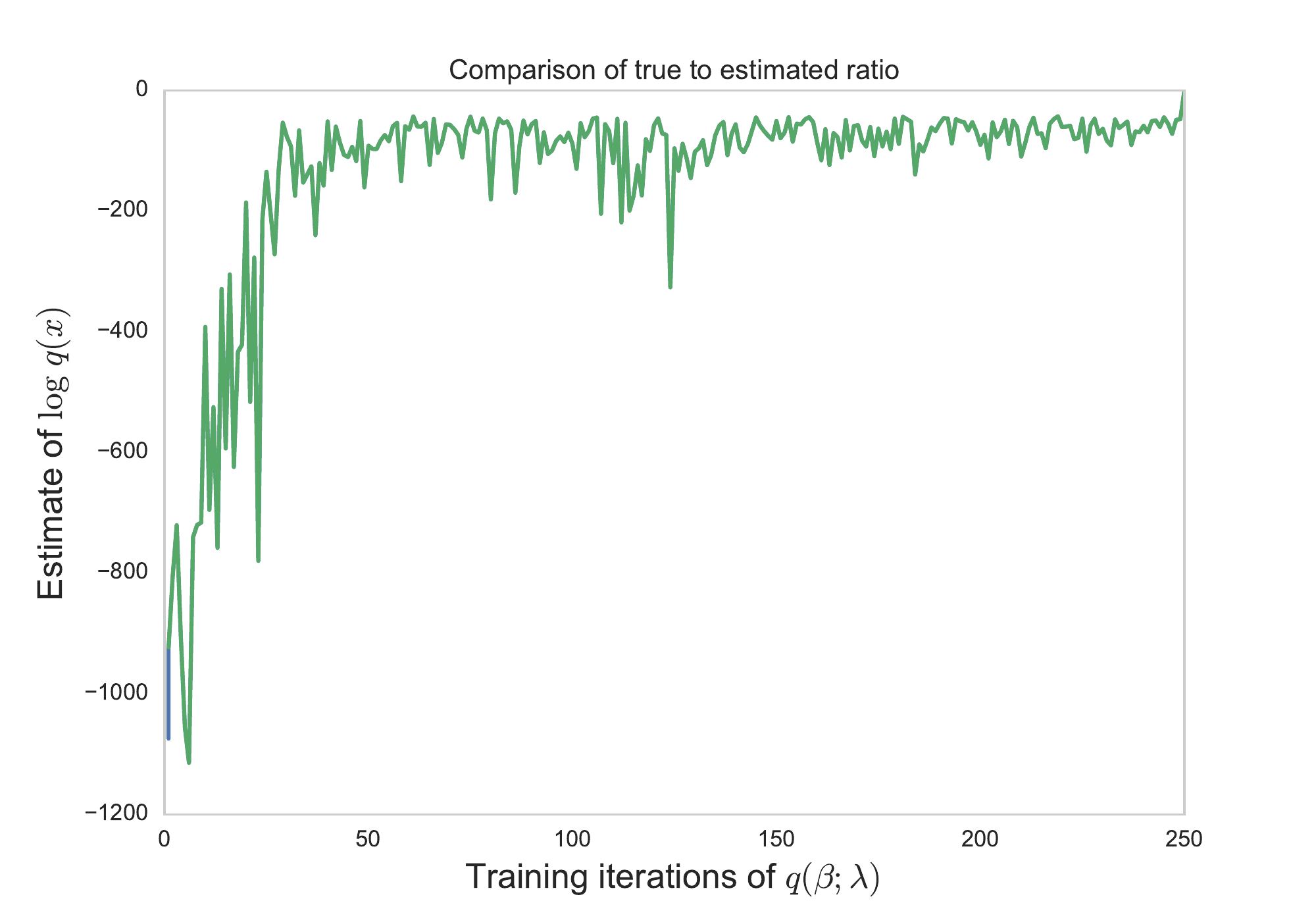}
\end{subfigure}%
\begin{subfigure}{0.33\columnwidth}
  \centering
  \includegraphics[width=\columnwidth]{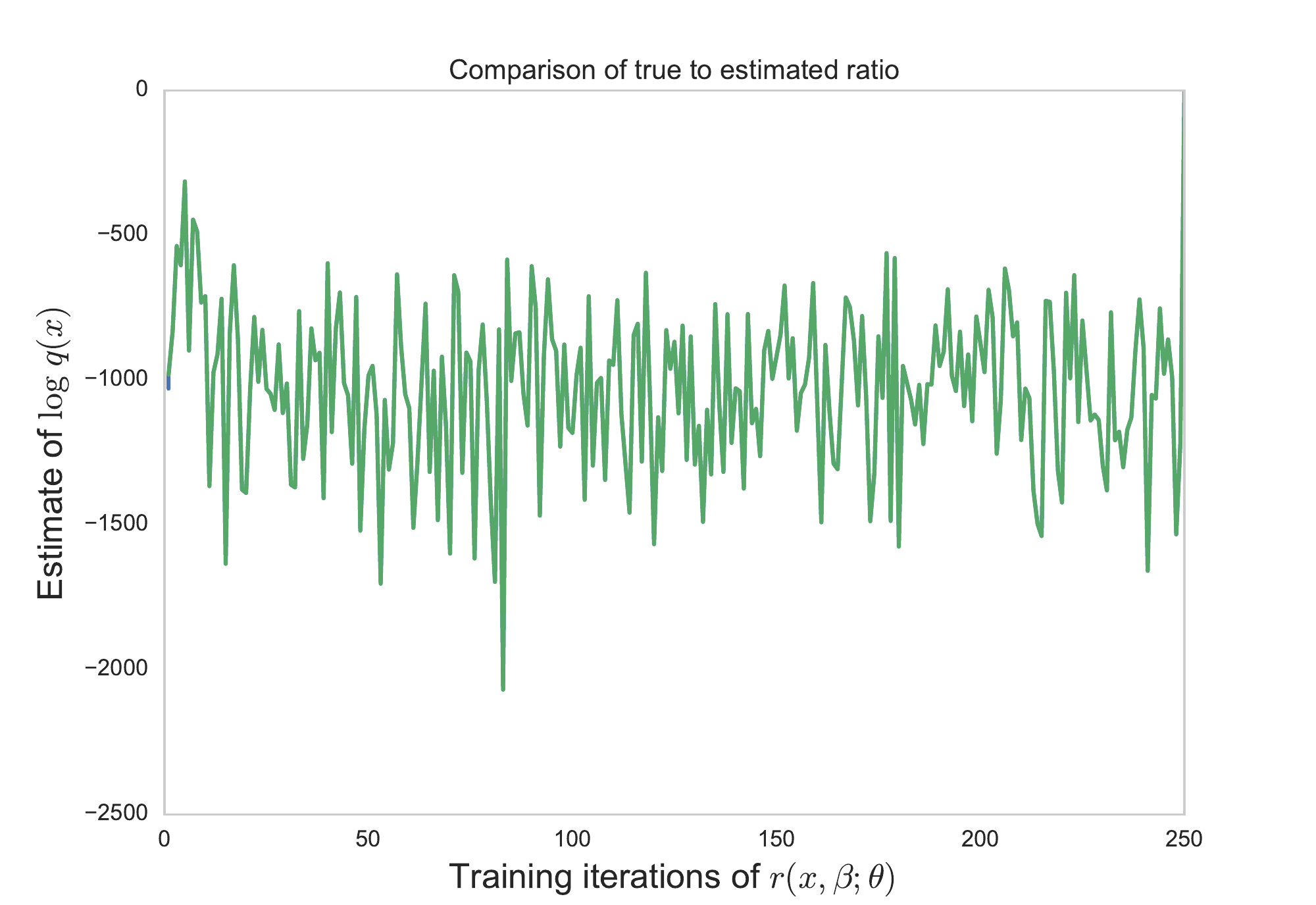}
\end{subfigure}%
\begin{subfigure}{0.33\columnwidth}
  \centering
  \includegraphics[width=\columnwidth]{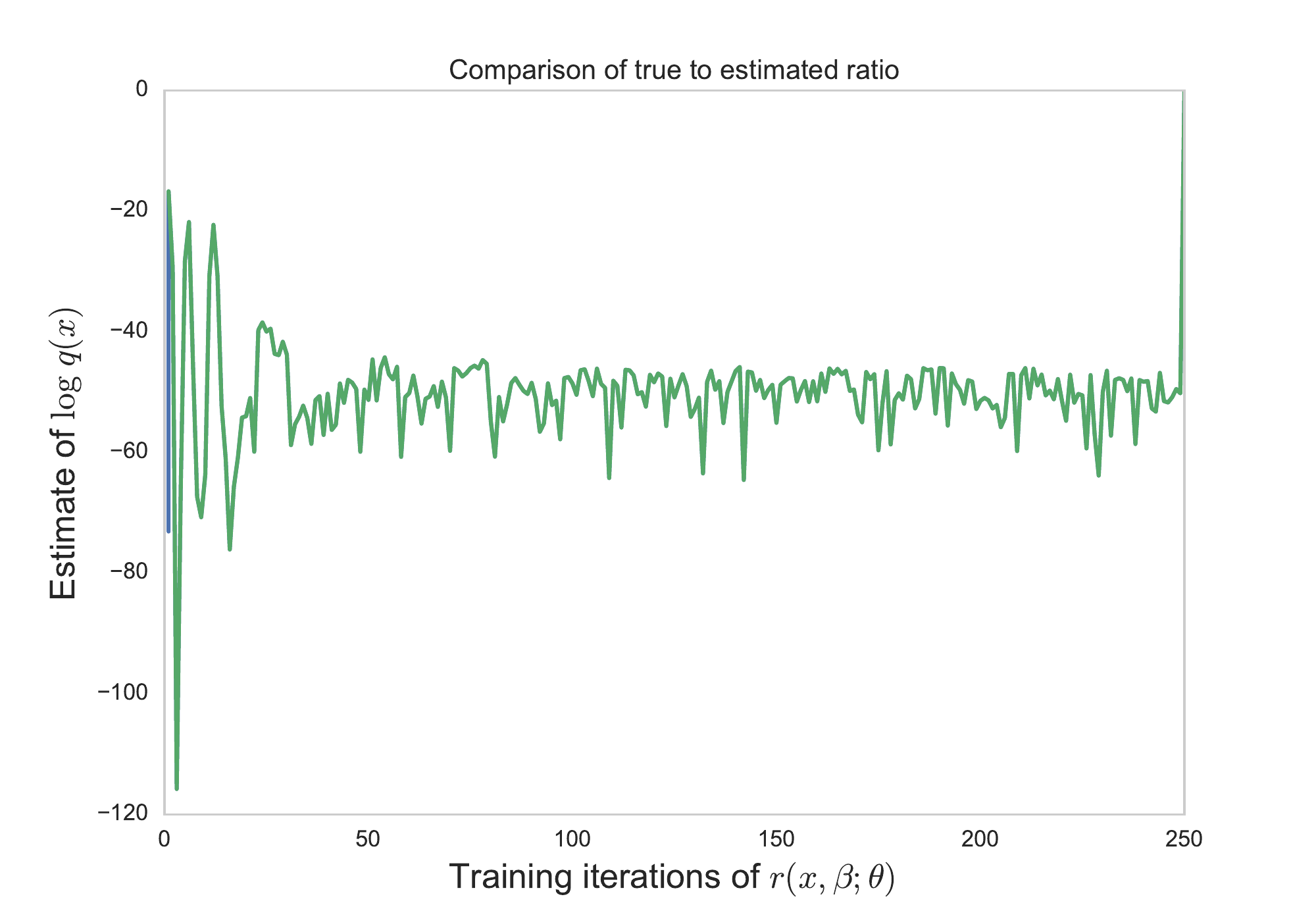}
\end{subfigure}
\caption{\textbf{(left)}
Difference of ratios over steps of $q$. Low
variance on $y$-axis means more stable. Interestingly, the ratio
estimator is more accurate and stable as $q$ converges to the posterior.
\textbf{(middle)}
Difference of ratios over steps of $r$;
$q$ is fixed at random initialization.
The ratio estimator doesn't improve even after many steps.
\textbf{(right)}
Difference of ratios over steps of $r$;
$q$ is fixed at the posterior.
The ratio estimator only requires few steps from random initialization to be
highly accurate.
}
\vspace{-1.25ex}
\label{fig:ratio_est}
\end{figure}

See \myfig{ratio_est}.
The top graph displays the estimate of $\log q(\mbx)$ over updates of the
variational approximation $q(\mbbeta)$; each estimate uses a sample from the
current $q(\mbbeta)$.
The ratio estimator $r$ is more accurate as $q$ exactly converges to
the posterior. This matches our intuition: if data generated from
the model is close to the true data, then the ratio is more stable
to estimate.

An alternative hypothesis for \myfig{ratio_est} is that the ratio
estimator has simply accumulated information during training. This
turns out to be untrue; see the bottom graphs.  On the left, $q$ is fixed at a
random initialization; the estimate of $\log q(\mbx)$ is displayed
over updates of $r$. After many updates, $r$ still produces unstable
estimates. In contrast, the right shows the same procedure with $q$
fixed at the posterior. $r$ is accurate after few updates.

Several practical insights appear for training.
First, it is not helpful to update $r$ multiple times before updating
$q$ (at least in initial iterations).
Additionally, if the specified model poorly matches the data, training
will be difficult across all iterations.

The property that ratio estimation is more accurate as the variational
approximation improves is because $q(\mbx_n)$ is set to be the
empirical distribution.
(Note we could subtract any density $q(\mbx_n)$ from the \gls{ELBO}
in
Equation 4.)
\Acrlong{LFVI}
finds $q(\mbbeta)$ that makes the observed data likely under
$p(\mbx_n \g \mbbeta)$, i.e., $p(\mbx_n \g \mbbeta)$
gets closer to the empirical distribution at values sampled from
$q(\mbbeta)$. Letting $q(\mbx_n)$ be the empirical distribution
means the ratio estimation problem will be less trivially solvable
(thus more accurate) as $q(\mbbeta)$ improves.

Note also this motivates why we do not subsume inference of
$p(\mbbeta\g\mbx)$ in the ratio in order to enable implicit global
variables and implicit global variational approximations. Namely,
estimation requires comparing samples between the prior and the
posterior; they rarely overlap for global variables.

\end{document}